\documentclass[sigconf,acmart]{acmart}

\usepackage{multirow}
\usepackage{makecell}
\usepackage{colortbl}
\definecolor{ourrow}{RGB}{234,242,252}
\graphicspath{{figures/}}

\AtBeginDocument{%
  }

\copyrightyear{2026}
\acmYear{2026}
\setcopyright{cc}
\setcctype{by}
\acmConference[DFF '26]{2nd Deepfake Forensics Workshop: Detection, Attribution, Recognition, and Adversarial Challenges in the Era of AI-Generated Media}{November 10--14, 2026}{Rio de Janeiro, Brazil}
\acmBooktitle{2nd Deepfake Forensics Workshop: Detection, Attribution, Recognition, and Adversarial Challenges in the Era of AI-Generated Media (DFF '26), November 10--14, 2026, Rio de Janeiro, Brazil}
\acmDOI{10.1145/3840473.3840520}
\acmISBN{979-8-4007-2927-0/2026/11}

\newcommand{\method}{AdaptPrompt}
\newcommand{\dataset}{Diff-Gen}

\begin{document}

\title{\method{}: Parameter-Efficient Adaptation of Vision-Language Models for Generalizable Deepfake Detection}

%%%%%%%%%%%%%%%%%%%% Authors (suppressed while [anonymous] is set) %%%%%%%%%%%%%%%%%%
\author{Yichen Jiang}
\authornote{Both authors contributed equally to this work. \textsuperscript{\textdagger}Corresponding author.}
\email{y92jiang@uwaterloo.ca}
\affiliation{%
  \institution{University of Waterloo}
  \city{Waterloo}
  \country{Canada}
}

\author{Mohammed Talha Alam\textsuperscript{*\textdagger}}
% \authornotemark[1]
% \authornote{Corresponding author.}
\email{mohammed.alam@mbzuai.ac.ae}
\orcid{0000-0001-5449-573X}
\affiliation{%
  \institution{MBZUAI}
  \city{Abu Dhabi}
  \country{UAE}
}

\author{Sohail Ahmed Khan}
\email{sohail.khan@uib.no}
\orcid{0000-0001-5351-2278}
\affiliation{%
  \institution{University of Bergen}
  \city{Bergen}
  \country{Norway}
}

\author{Duc-Tien Dang-Nguyen}
\email{ductien.dangnguyen@uib.no}
\orcid{0000-0002-2761-2213}
\affiliation{%
  \institution{University of Bergen}
  \city{Bergen}
  \country{Norway}
}

\author{Fakhri Karray}
\email{karray@uwaterloo.ca}
\orcid{0000-0002-6900-315X}
\affiliation{%
  \institution{University of Waterloo}
  \city{Waterloo}
  \country{Canada}
}

\renewcommand{\shortauthors}{Jiang et al.}

\begin{abstract}
Detectors of AI-generated images tend to inherit the biases of the data they are trained on: models fitted to GAN imagery learn to treat GAN-specific artifacts as the very definition of ``fake'' and consequently miss images produced by diffusion models and commercial generation tools. We study this generalization problem from two directions. First, we introduce \dataset{}, a balanced corpus consisting of 100k diffusion-generated samples and an equally sized set of real images, with the real subset selected to mirror the class distribution of the synthetic data. A spectral analysis shows that, unlike GAN data, \dataset{} exhibits broad, non-periodic high-frequency energy, and we find that detectors trained on it transfer substantially better to unseen generator families. Second, we propose \method{}, a parameter-efficient adaptation of CLIP that combines a visual adapter with learnable text prompts and trains roughly 0.1\% of the model's parameters. We further observe that truncating the last transformer block of the vision encoder consistently improves detection, suggesting that the final semantic-alignment layers of CLIP suppress the low-level traces on which forensic decisions rely. Across a benchmark of 25 test sets covering GANs, diffusion models, and commercial tools such as Midjourney and DALL-E~3, \method{} trained on \dataset{} attains the best mean average precision (98.60\%) and accuracy (92.72\%), while matching fully fine-tuned baselines at a fraction of their training cost. We also show that the same framework supports data-efficient training and closed-set source attribution across 22 generators.
\end{abstract}




\maketitle

\section{Introduction}
Modern generative models can produce images that are essentially indistinguishable from photographs, and the resulting risks, including disinformation, impersonation, fraud, and the erosion of trust in visual evidence, are well documented~\cite{bird2023typologyrisksgenerativetexttoimage, chen2025prevalencesharingpatternsspreaders}. Safeguards from safety alignment of text-to-image models~\cite{alam2025spqr} to privacy-preserving uses of face synthesis~\cite{alam2025faceanonymixer}, address part of the problem, but for content already in circulation the burden falls on detection, which has become a central problem in multimedia forensics. The difficulty is not detecting images from a known generator, which is largely solved, but generalizing to generators never seen during training~\cite{wang2020cnn, khan2023deepfake, zhu2021faceforgery, chen2022self}.

This difficulty has grown with the shift from Generative Adversarial Networks (GANs)~\cite{goodfellow2021GAN} to diffusion models such as Stable Diffusion and Midjourney~\cite{rombach2022latentdiffusion, midjourney2023}. Detectors trained on GAN imagery rely, often implicitly, on the periodic upsampling artifacts that GANs leave in the frequency domain~\cite{durall2020watch, frank2020leveraging}. Diffusion models do not produce these patterns, and \citet{ojha2023universal} showed the consequence: a binary classifier that has learned GAN fingerprints as its notion of ``fake'' assigns everything else, including diffusion output, to the ``real'' class, which acts as a sink label. Two ingredients of the detection pipeline are implicated at once: the training data, which fixes the artifacts a model can learn, and the adaptation strategy, which determines how much of a pre-trained representation survives fine-tuning.

\begin{figure}[t]
    \centering
    \includegraphics[width=\linewidth]{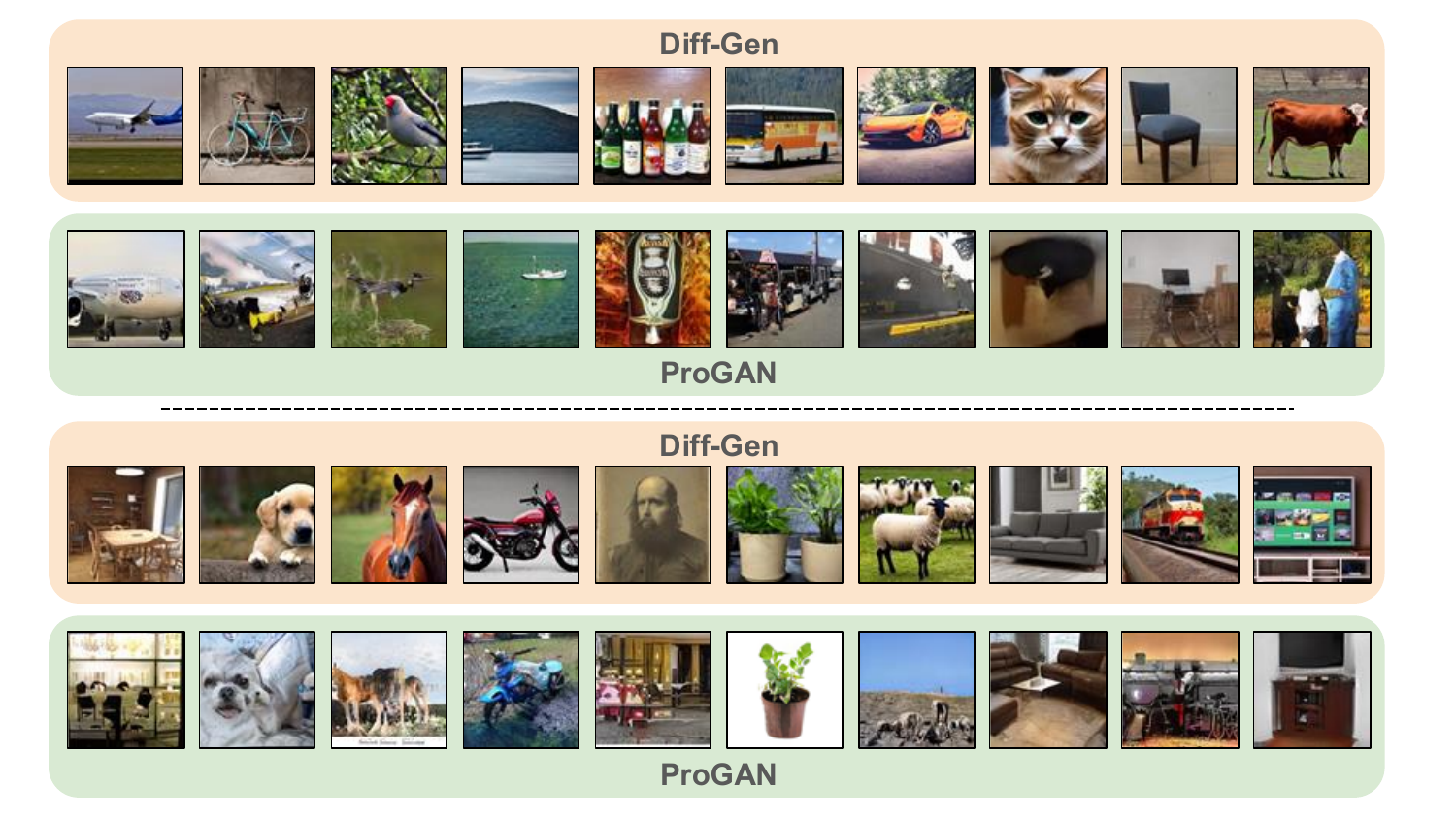}
    \caption{Random samples from \dataset{} (diffusion-generated) and from the ProGAN training set of \citet{wang2020cnn}. The two corpora share the same 20 LSUN object categories and class distribution, so a detector trained on either sees the same semantic content and differs only in the generative process that produced the fakes.}
    \label{fig:dataset}
\end{figure}

We address both aspects in this paper. From the data perspective, we introduce \dataset{}, a training corpus comprising 100k diffusion-generated images whose class distribution is aligned with that of 100k real LSUN images~\cite{yu2015lsun} (Fig.~\ref{fig:dataset}). Its power spectrum shows the broad, non-periodic high-frequency elevation characteristic of diffusion synthesis rather than the sharp spectral peaks of GAN data, and we ask a question that, to our knowledge, has not been answered systematically: does training on diffusion artifacts generalize \emph{backward} to GANs and \emph{forward} to commercial tools better than the reverse? On the modeling side, we build on the observation that CLIP features transfer remarkably well to forensic tasks~\cite{radford2021learning, ojha2023universal, khan2024clipping} and propose \method{}, which adapts both pathways of CLIP at once, using a residual adapter on the visual side~\cite{gao2023clipadapter} and learnable prompt vectors on the textual side~\cite{zhou2022learning}, while the backbone itself stays frozen. Only about 602k parameters, roughly 0.1\% of the model, are trained. Our contributions are as follows.
\begin{itemize}
    \item We introduce \dataset{}, a diffusion-based training dataset for deepfake detection, and show that models trained on it generalize markedly better to diffusion and commercial generators than models trained on the widely used ProGAN corpus~\cite{karras2018progressive}, while remaining competitive on GANs.
    \item We propose \method{}, a dual-modality parameter-efficient adaptation of CLIP, and show that truncating the final transformer block of the vision encoder consistently improves forensic performance, providing evidence that CLIP's last semantic-alignment layers discard the low-level traces detection depends on.
    \item We evaluate on 25 test sets spanning ten GANs, nine diffusion models, three commercial tools, and face-swap benchmarks, reporting the best mean average precision and accuracy among all compared methods, together with analyses of robustness, data efficiency, dataset bias, and source attribution over 22 generators.
\end{itemize}

\section{Related Work}

\subsection{Detecting Generated Images}
Early work on synthetic-image forensics targeted manipulations of real photographs, such as splicing and face editing~\cite{wang2019detecting, perov2020deepfacelab, rossler2019faceforensics}, or visible physical inconsistencies in lighting and perspective~\cite{farid2022lighting, farid2022perspective, matern2019exploiting}. With the rise of GANs, attention moved to statistical traces of the generation process itself: GAN images carry characteristic periodic patterns in the frequency domain, introduced by upsampling layers, that make them separable from camera output~\cite{durall2020watch, dzanic2020fourier, frank2020leveraging}. Such joint image-spectral cues have proven effective well outside natural photography, for instance in detecting synthetic astronomical imagery~\cite{alam2024astrospy}. \citet{wang2020cnn} showed that a ResNet-50~\cite{he2016deepresidual} trained on 720k images (half real LSUN~\cite{yu2015lsun}, half ProGAN~\cite{karras2018progressive}) generalizes surprisingly well, but only within the GAN family. Diffusion models~\cite{sohl2015deep, rombach2022latentdiffusion} synthesize images by iterative denoising and leave residuals that resemble broadband noise rather than periodic structure, and several studies found that GAN-trained detectors degrade sharply on them~\cite{gragnaniello2021gan, corvi2023detection}. This behavior arises from the sink-label effect: once a classifier learns to associate the ``fake'' class with a particular family of artifacts, samples that do not exhibit those cues are by default classified as ``real''~\cite{ojha2023universal}.

\subsection{Adapting Vision-Language Models for Forensics}
A productive response to the sink-label problem is to start from a feature space that was not learned on any generator at all. \citet{ojha2023universal} showed that a linear classifier over frozen CLIP-ViT features~\cite{radford2021learning} generalizes across GAN and diffusion families better than end-to-end trained CNNs; the transferability of CLIP representations to distant domains, such as astronomical imaging~\cite{imam2024cosmoclip}, makes them a natural backbone for forensic tasks as well. \citet{khan2024clipping} then compared adaptation strategies for CLIP-based detection, including linear probing, full fine-tuning, adapter networks~\cite{gao2023clipadapter}, and prompt tuning~\cite{zhou2022learning}, across 21 test sets, and found that the two parameter-efficient methods generalize best, whereas full fine-tuning tends to overfit the training generator; similar benefits of careful transfer-learning design under domain shift have been observed in other imbalanced imaging problems~\cite{imam2023optimizing}. Beyond raw detection scores, robustness and calibration of authenticity classifiers are increasingly seen as requirements in their own right~\cite{hashmi2025robustcalibrateddetectionauthentic}. Relatedly, \citet{cozzolino2023raising} reported that features drawn from earlier layers of CLIP can be more discriminative for forensic tasks than the final output features. These observations leave a natural question open: adapters act only on the visual pathway and prompt tuning only on the textual one, so what is gained by adapting both jointly, and how much of the forensic signal is lost in CLIP's final layers? \method{} is our answer to both questions.

\begin{figure*}[t]
    \centering
    \includegraphics[width=0.75\textwidth]{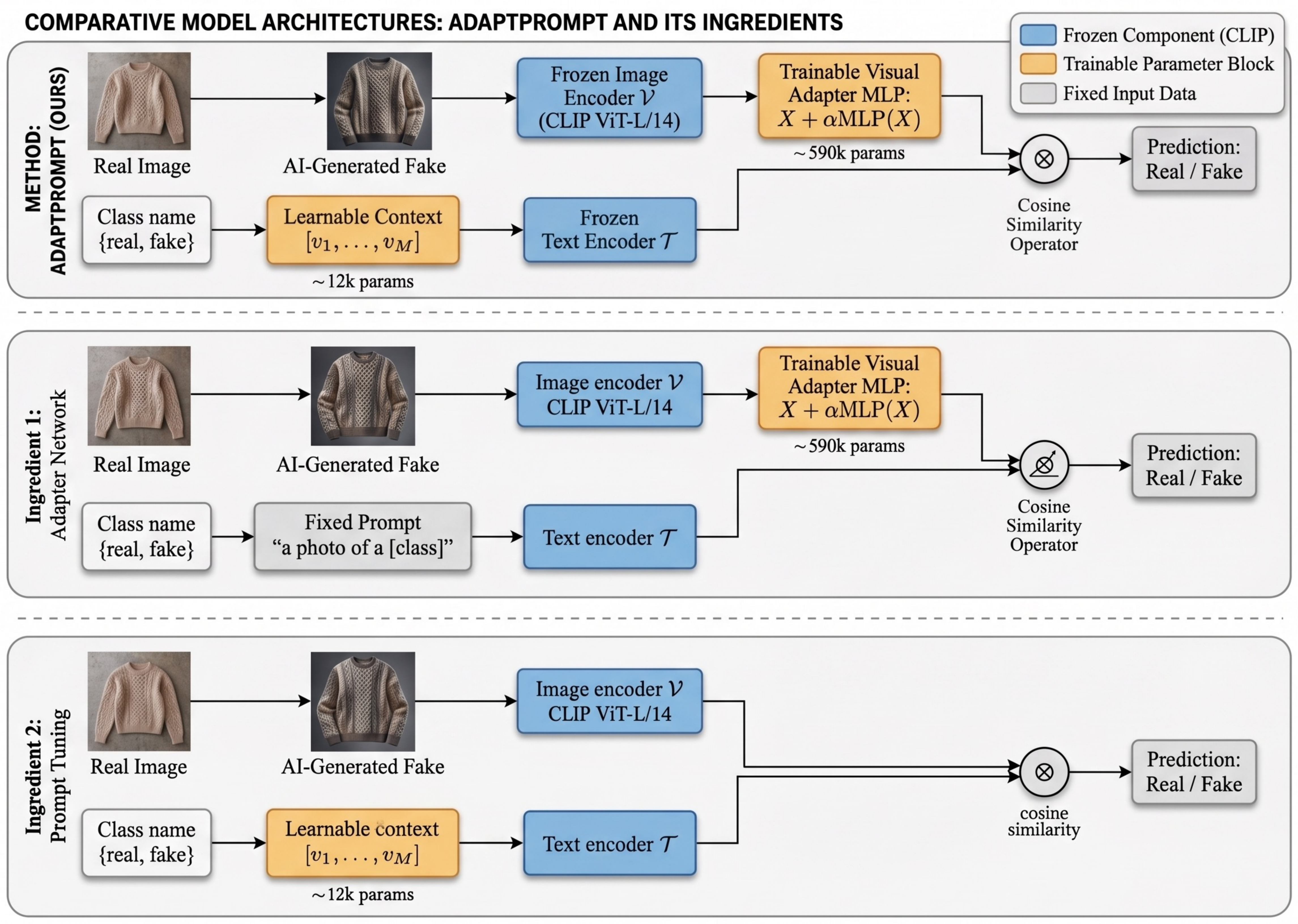}
    \caption{Overview of \method{} (top) next to its two ingredients, the adapter network (middle) and prompt tuning (bottom). Blue blocks are the frozen CLIP encoders, orange blocks are trainable, and gray blocks are fixed inputs. \method{} trains a residual adapter on the visual pathway (${\sim}590$k parameters) and learnable context vectors on the textual pathway (${\sim}12$k parameters) under a single classification loss; the prediction is the cosine similarity between adapted image features and class embeddings, and all CLIP weights stay fixed.}
    \label{fig:method}
\end{figure*}

\section{Method}

\subsection{Problem Setting}
Deepfake detection is binary classification: learn $f:\mathcal{X}\rightarrow\{0,1\}$ mapping an image $x$ to $y=1$ (fake) or $y=0$ (real). The practical difficulty is distribution shift on the fake class. A detector is trained on fakes from a source family of generators $\mathcal{D}_S$ but deployed against fakes from unseen families $\mathcal{D}_T$ whose artifacts differ in both spatial and spectral structure. Throughout, all evaluation is cross-generator: no test-set generator is ever seen during training unless explicitly noted.

\subsection{Background: CLIP-Based Detection}
CLIP~\cite{radford2021learning} pairs a visual encoder $\mathcal{V}$ and a text encoder $\mathcal{T}$ that project images and captions into a shared embedding space. Prior forensic work freezes $\mathcal{V}$ and trains a lightweight classifier on the embeddings $z_v=\mathcal{V}(x)$~\cite{ojha2023universal, khan2024clipping}. This preserves the generality of the pre-trained features but treats them as fixed: nothing encourages the representation to retain the faint, non-semantic residuals that distinguish generated images. We use the ViT-L/14 backbone~\cite{dosovitskiy2021vit} (427M parameters, 768-dimensional joint embedding space) in all experiments, following~\cite{ojha2023universal, khan2024clipping}.

\subsection{\method{}}
\method{} adapts both CLIP pathways simultaneously with two small trainable modules (Fig.~\ref{fig:method}).

\subsubsection{\textbf{Visual adapter:}}
On the visual side we attach a residual bottleneck adapter~\cite{gao2023clipadapter} to the output of the frozen encoder. For image features $X\in\mathbb{R}^{B\times d_v}$,
\begin{equation}
    Y = X + \alpha\cdot\mathrm{MLP}(X),
    \qquad
    \mathrm{MLP}(X)=\sigma(XW_{\mathrm{down}})\,W_{\mathrm{up}},
\end{equation}
where $W_{\mathrm{down}}\in\mathbb{R}^{d_v\times d_{\mathrm{mid}}}$, $W_{\mathrm{up}}\in\mathbb{R}^{d_{\mathrm{mid}}\times d_v}$, $\sigma$ is a ReLU, and $\alpha$ is a scaling factor. We use $d_v=768$ and $d_{\mathrm{mid}}=384$, giving roughly 590k trainable parameters. The bottleneck forces the module to encode a compact correction to the semantic CLIP features, specifically the artifact-sensitive directions that the frozen representation carries but does not emphasize.

\subsubsection{\textbf{Textual prompt tuning:}}
For the textual branch, we move beyond manually specified prompts such as ``a photo of a fake'' by learning the prompt representations directly during training~\cite{zhou2022learning}. The prompt for class $c\in\{\textit{real},\textit{fake}\}$ is
\begin{equation}
    P_c = [v_1, v_2, \ldots, v_M, \mathrm{CLASS}_c],
\end{equation}
where the $v_m\in\mathbb{R}^{d}$ are $M=16$ shared learnable context vectors (${\sim}12$k parameters) and $\mathrm{CLASS}_c$ is the fixed embedding of the class name. The frozen text encoder maps each prompt to a class embedding $E_c=\mathcal{T}(P_c)$.

\subsubsection{\textbf{Training objective:}}
Classification uses the cosine similarity between adapted image features and class embeddings,
\begin{equation}
    p(y=c \mid x) =
    \frac{\exp\left(\langle Y, E_c\rangle/\tau\right)}
         {\sum_{j\in\{\textit{real},\textit{fake}\}}\exp\left(\langle Y, E_j\rangle/\tau\right)},
\end{equation}
with temperature $\tau$, and both modules are optimized jointly with cross-entropy. In total, \method{} trains about 602k parameters, roughly 0.1\% of the 427M-parameter backbone, so a training run is inexpensive and the pre-trained representation cannot be catastrophically overwritten. At inference, AdaptPrompt relies solely on the adapted visual embedding and learned real/fake text prompts, without generator-specific calibration, post-processing, or access to the source model.

\subsubsection{\textbf{Truncating the vision encoder:}}
CLIP's final layers are optimized to align image features with text, an objective that rewards semantic abstraction and suppresses the low-level irregularities a forensic model needs. Motivated by this, and by the layer analysis of \citet{cozzolino2023raising}, we consider three variants of every model: \textbf{v0} uses the full backbone; \textbf{v1} removes the final projection layer mapping visual features into the joint text-image space; \textbf{v2} additionally removes the last transformer block, so the adapter reads the penultimate block's features. If the final layers were forensically neutral, v1 and v2 should perform no better than v0; as Section~\ref{sec:ablation} shows, v2 is consistently strongest, and pre-last-block features retain measurably more high-frequency detail. One caveat: with plain linear probing (no adapter), v1 failed to converge in our experiments, indicating that pre-projection features are informative but not linearly separable; the non-linear adapter makes the truncated backbone usable.

\section{The \dataset{} Dataset}
\label{sec:diffgen}
Almost all large-scale training corpora for deepfake detection are GAN-based, the 720k-image ProGAN/LSUN set of \citet{wang2020cnn} being the de facto standard. \dataset{} is its diffusion-era counterpart: 100k fake images produced with Stable Diffusion v1.5~\cite{rombach2022latentdiffusion} and 100k real LSUN images with an identical distribution over the same 20 object categories (airplane, bird, bottle, and so on; Fig.~\ref{fig:dataset}). Images are generated from prompts of the form \emph{``A photo of a [class name]''} with a classifier-free guidance scale of 7.5, which balances fidelity against diversity, a trade-off known to govern how useful a synthetic corpus is for downstream training, both for augmentation~\cite{alam2024flare} and for the diversity of the resulting data~\cite{alam2024introducing}. Matching the class distribution of the real set is deliberate: it prevents a detector from exploiting semantic differences between the real and fake classes, leaving the generative artifacts as the only reliable signal.

\begin{figure}[t]
    \centering
    \includegraphics[width=0.92\linewidth]{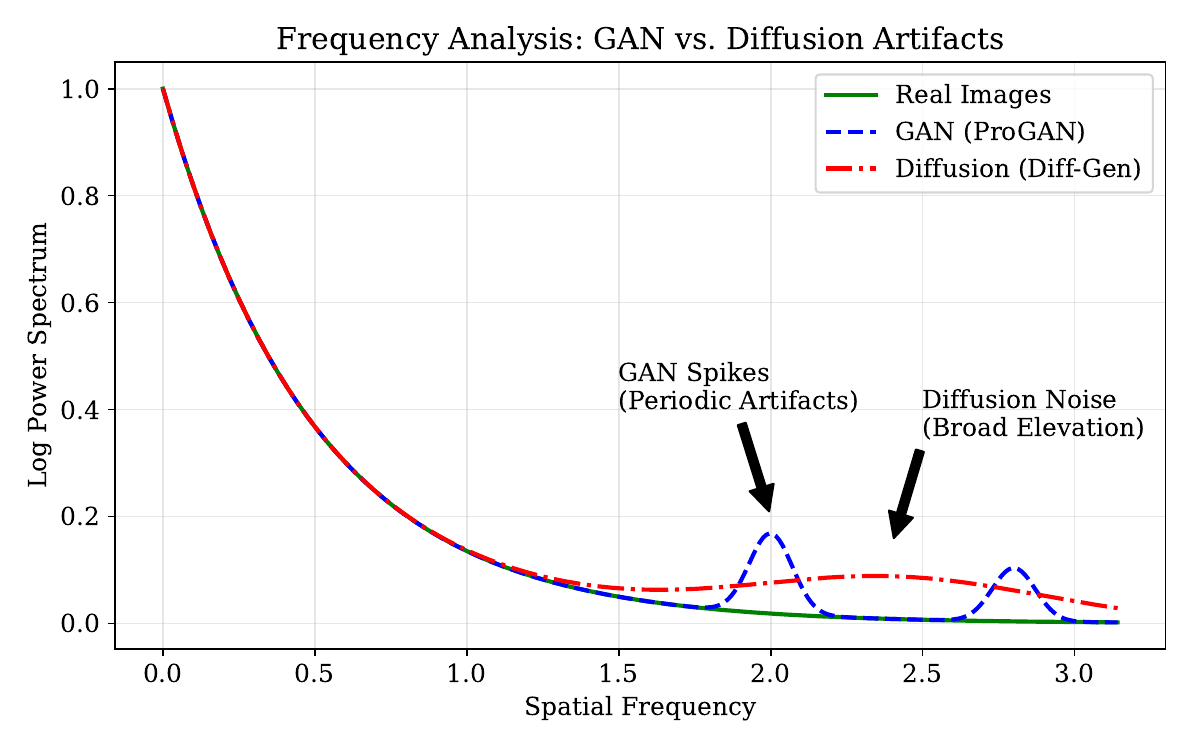}
    \caption{Azimuthally integrated power spectra of the training corpora. ProGAN (blue, dashed) shows the sharp high-frequency peaks caused by periodic upsampling artifacts; \dataset{} (red, dash-dot) instead shows a broad elevation over the real-image spectrum (green), reflecting the non-periodic noise residuals of diffusion synthesis.}
    \label{fig:spectral}
\end{figure}

Figure~\ref{fig:spectral} explains why the choice of training corpus matters so much. The ProGAN spectrum contains sharp, isolated peaks at high frequencies, corresponding to the periodic ``checkerboard'' signature of transposed convolutions. A detector can reach perfect training accuracy by keying on those peaks alone, which is precisely the shortcut that fails on diffusion images. \dataset{}, by contrast, deviates from the real spectrum through a broad, continuous elevation of high-frequency energy. There is no narrow band to latch onto, so a detector trained on \dataset{} is pushed toward decision boundaries that describe generated imagery in general rather than one architecture's fingerprint. The cross-generator results in Section~\ref{sec:results} bear this out.

\section{Experiments}
\label{sec:results}

\subsection{Setup}
\subsubsection{\textbf{Training corpora:}}
Following the protocol of \citet{khan2024clipping}, every method is trained either on the ProGAN corpus of \citet{wang2020cnn} (720k images) or on \dataset{} (200k images), and never sees any test generator during training.

\subsubsection{\textbf{Test benchmark:}}
We evaluate on 25 test sets in three groups: ten GANs (ProGAN, BigGAN, CycleGAN, EG3D, GauGAN, StarGAN, StyleGAN~1--3, Taming Transformers); nine diffusion models (three Glide configurations, Guided Diffusion, three LDM configurations, Stable Diffusion, SDXL), whose per-model scores we report averaged over their configurations; three commercial tools (Midjourney~v5, Adobe Firefly, DALL-E~3); and, additionally, DALL-E~mini and the DeepFakes and FaceSwap subsets of FaceForensics++~\cite{rossler2019faceforensics}. Full per-set statistics, real-image sources, and resolutions are given in Table~\ref{table:datasets}. We report average precision (AP) and classification accuracy (Acc) with a fixed 0.5 threshold, computed with scikit-learn~\cite{scikit-learn-docs}.

\subsubsection{\textbf{Baselines:}}
We compare against the CNN detectors of \citet{wang2020cnn}, \citet{gragnaniello2021gan}, and \citet{corvi2023detection}; the CLIP linear probe of \citet{ojha2023universal}; and the four CLIP adaptation strategies studied by \citet{khan2024clipping} (linear probing, full fine-tuning, adapter, prompt tuning), all trained on ProGAN. We then retrain those same four strategies on \dataset{}, in all three encoder variants (v0/v1/v2), alongside \method{}. This factorial design separates the effect of the training data from the effect of the adaptation method.

\begin{table*}[t]
  \centering
  \caption{Average precision on unseen test sets (all values \%). Rows above the rule are trained on ProGAN; the \dataset{} blocks are trained on our dataset and correspond to the full (v0), projection-removed (v1), and last-block-removed (v2) encoders. This suite is dominated by GAN test sets, which favors ProGAN-trained models; the picture across all three generator families, including commercial tools, is summarized in Table~\ref{tab:main}. Glide and LDM scores are averaged over their three configurations each. Bold marks columns where a \dataset{}-trained model is best overall; shaded rows are \method{}.}
  \label{tab:allAP}
  \resizebox{\textwidth}{!}{%
    \begin{tabular}{llccccccccccccccccccc}
    \toprule
    \multirow{2}{*}{Training} & \multirow{2}{*}{Method}
      & \multicolumn{10}{c}{Generative adversarial networks}
      & \multirow{2}{*}{\makecell{DALL-E\\(mini)}}
      & \multicolumn{5}{c}{Diffusion models}
      & \multicolumn{2}{c}{FaceForensics++}
      & \multirow{2}{*}{mAP} \\
    \cmidrule(lr){3-12} \cmidrule(lr){14-18} \cmidrule(lr){19-20}
      & & \makecell{Pro\\GAN} & \makecell{Big\\GAN} & \makecell{Cycle\\GAN} & EG3D & \makecell{Gau\\GAN} & \makecell{Star\\GAN}
      & \makecell{Style\\GAN} & \makecell{Style\\GAN2} & \makecell{Style\\GAN3} & \makecell{Taming\\Transf.}
      & & Glide & Guided & LDM & SD & SDXL
      & \makecell{Deep\\Fakes} & \makecell{Face\\Swap} & \\
    \midrule
    \multirow{10}{*}{ProGAN}
      & Wang et al.~\cite{wang2020cnn} (B+J 0.1)
      & 100.00 & 83.04 & 90.09 & 95.58 & 88.94 & 97.18 & 99.27 & 96.43 & 98.63 & 73.90
      & 67.47 & 81.02 & 83.10 & 68.61 & 64.33 & 72.27 & 75.88 & 50.78 & 81.18 \\
      & Wang et al.~\cite{wang2020cnn} (B+J 0.5)
      & 100.00 & 82.63 & 94.71 & 55.32 & 96.62 & 93.88 & 93.25 & 88.64 & 85.33 & 59.78
      & 60.92 & 69.75 & 65.11 & 60.24 & 52.14 & 65.92 & 64.33 & 49.76 & 72.65 \\
      & Gragnaniello et al.~\cite{gragnaniello2021gan}
      & 100.00 & 97.57 & 97.63 & 99.95 & 98.36 & 99.99 & 100.00 & 99.98 & 100.00 & 95.31
      & 91.32 & 94.08 & 93.81 & 92.33 & 91.75 & 90.93 & 95.90 & 61.54 & 94.24 \\
      & Corvi et al.~\cite{corvi2023detection} (ProGAN)
      & 100.00 & 99.66 & 97.94 & 99.92 & 99.74 & 99.95 & 100.00 & 99.96 & 99.93 & 94.34
      & 95.45 & 89.51 & 79.30 & 88.26 & 87.01 & 74.90 & 95.52 & 56.58 & 91.52 \\
      & Corvi et al.~\cite{corvi2023detection} (Latent)
      & 91.83 & 74.25 & 49.05 & 42.87 & 89.14 & 50.19 & 73.25 & 74.73 & 70.20 & 95.21
      & 98.15 & 87.35 & 59.17 & 100.00 & 100.00 & 99.23 & 83.70 & 45.52 & 79.93 \\
      & Ojha et al.~\cite{ojha2023universal}
      & 99.99 & 98.73 & 98.92 & 79.58 & 99.74 & 96.06 & 95.73 & 95.81 & 92.21 & 97.12
      & 96.84 & 93.85 & 92.09 & 95.71 & 93.58 & 88.55 & 77.48 & 75.87 & 93.05 \\
      & Linear probing~\cite{khan2024clipping}
      & 99.91 & 97.77 & 98.53 & 99.48 & 99.69 & 99.00 & 95.53 & 94.98 & 99.54 & 97.74
      & 95.65 & 97.75 & 92.14 & 95.94 & 92.24 & 94.99 & 80.07 & 76.58 & 95.22 \\
      & Fine-tuning~\cite{khan2024clipping}
      & 100.00 & 98.65 & 99.00 & 99.97 & 98.12 & 100.00 & 99.61 & 99.48 & 100.00 & 98.38
      & 98.15 & 96.23 & 97.40 & 98.79 & 97.53 & 99.52 & 87.42 & 60.22 & 96.29 \\
      & Adapter~\cite{khan2024clipping}
      & 100.00 & 99.58 & 99.97 & 99.50 & 99.98 & 99.98 & 99.44 & 98.80 & 99.83 & 99.27
      & 98.60 & 99.26 & 96.16 & 97.76 & 91.90 & 92.32 & 91.37 & 82.11 & 97.27 \\
      & Prompt tuning~\cite{khan2024clipping}
      & 100.00 & 99.42 & 99.92 & 99.51 & 99.95 & 99.97 & 99.52 & 98.62 & 99.68 & 99.54
      & 98.89 & 99.32 & 97.41 & 97.91 & 96.23 & 96.42 & 92.59 & 88.01 & 98.06 \\
    \midrule
    \multirow{5}{*}{\makecell[l]{\dataset{}\\(v0)}}
      & Linear probing
      & 97.64 & 74.51 & 71.70 & 98.00 & 71.56 & 89.04 & 99.18 & 95.14 & 99.72 & 99.22
      & 97.98 & 83.45 & 96.19 & 92.97 & 99.89 & 99.83 & 47.65 & 55.70 & 87.50 \\
      & Fine-tuning
      & 99.90 & 70.93 & 60.61 & \textbf{100.00} & 61.54 & 99.72 & 98.76 & 98.79 & \textbf{100.00} & 99.62
      & \textbf{99.81} & 95.89 & 97.93 & 99.86 & 99.97 & 99.95 & 63.09 & 55.57 & 91.73 \\
      & Adapter
      & 99.90 & 94.87 & 91.51 & 99.97 & 95.41 & 99.10 & 99.20 & 97.51 & 99.95 & 99.54
      & 99.43 & 97.01 & 97.40 & 98.01 & 99.96 & 99.80 & 66.17 & 66.39 & 95.53 \\
      & Prompt tuning
      & 99.79 & 88.53 & 79.81 & 99.97 & 85.35 & 98.46 & 99.29 & 97.08 & 99.81 & 99.32
      & 99.37 & 95.22 & 96.59 & 97.26 & 99.86 & 99.71 & 61.07 & 61.30 & 93.62 \\
      \rowcolor{ourrow}
      & \method{}
      & 99.87 & 91.53 & 89.59 & 99.95 & 89.69 & 98.69 & 98.91 & 96.84 & 99.72 & 99.09
      & 99.08 & 95.15 & 96.04 & 97.71 & 99.98 & 99.73 & 63.95 & 65.01 & 94.50 \\
    \midrule
    \multirow{4}{*}{\makecell[l]{\dataset{}\\(v1)}}
      & Fine-tuning
      & 99.41 & 65.50 & 52.75 & \textbf{100.00} & 65.88 & 99.48 & 97.93 & 97.96 & \textbf{100.00} & 99.52
      & 99.16 & 93.43 & 97.23 & 99.71 & \textbf{100.00} & \textbf{99.98} & 54.37 & 52.73 & 90.44 \\
      & Adapter
      & 99.90 & 94.86 & 91.50 & 99.97 & 95.41 & 99.09 & 99.20 & 97.51 & 99.95 & 99.54
      & 99.44 & 97.01 & 97.39 & 98.01 & 99.96 & 99.80 & 66.16 & 66.37 & 95.53 \\
      & Prompt tuning
      & 99.79 & 88.53 & 79.81 & 99.97 & 85.35 & 98.46 & 99.29 & 97.08 & 99.81 & 99.32
      & 99.37 & 95.22 & 96.59 & 97.26 & 99.86 & 99.71 & 61.07 & 61.30 & 93.62 \\
      \rowcolor{ourrow}
      & \method{}
      & 99.87 & 91.53 & 89.59 & 99.95 & 89.69 & 98.69 & 98.91 & 96.84 & 99.72 & 99.09
      & 99.08 & 95.15 & 96.04 & 97.38 & 99.98 & 99.73 & 63.95 & 65.01 & 94.46 \\
    \midrule
    \multirow{5}{*}{\makecell[l]{\dataset{}\\(v2)}}
      & Linear probing
      & 98.53 & 83.06 & 75.70 & 96.59 & 79.07 & 95.88 & 98.99 & 95.01 & 99.56 & 99.25
      & \textbf{99.81} & 85.50 & 95.98 & 92.76 & 99.85 & 99.78 & 63.09 & 55.57 & 89.81 \\
      & Fine-tuning
      & 99.77 & 70.84 & 59.23 & \textbf{100.00} & 62.49 & 99.02 & 97.87 & 98.50 & \textbf{100.00} & 99.56
      & 99.55 & 95.77 & 97.35 & 99.75 & 99.99 & 99.94 & 62.06 & 53.54 & 91.43 \\
      & Adapter
      & 99.76 & 95.64 & 95.93 & 99.99 & 97.00 & 99.76 & 98.89 & 96.26 & 99.89 & 99.50
      & 99.23 & 96.70 & 97.52 & 97.74 & 99.99 & 99.91 & 69.66 & 69.35 & 96.02 \\
      & Prompt tuning
      & 99.78 & 90.29 & 89.93 & 99.85 & 90.70 & 99.80 & 99.17 & 95.91 & 99.85 & 98.77
      & 98.47 & 95.10 & 95.56 & 96.72 & 99.74 & 99.52 & 55.98 & 60.32 & 93.88 \\
      \rowcolor{ourrow}
      & \method{}
      & 99.82 & 94.24 & 95.46 & 99.99 & 95.50 & 99.78 & 99.01 & 96.41 & 99.95 & \textbf{99.68}
      & 99.33 & 97.66 & \textbf{98.24} & 97.95 & 99.98 & 99.90 & 67.06 & 68.17 & 95.91 \\
    \bottomrule
    \end{tabular}%
  }
\end{table*}

\begin{table}[t]
    \centering
    \caption{Mean AP and accuracy (\%) per generator family over the full benchmark, including the commercial tools. Bold marks columns where a \dataset{}-trained model is best overall; shaded rows are \method{}. Per-dataset accuracies are reported in Appendix~\ref{sec:appacc} (Table~\ref{tab:allAcc}).}
    \label{tab:main}
    \resizebox{\linewidth}{!}{%
    \begin{tabular}{llcccc}
    \toprule
    Training & Method & GAN & Diffusion & Comm.\ tools & Average \\
     & & AP\,/\,Acc & AP\,/\,Acc & AP\,/\,Acc & AP\,/\,Acc \\
    \midrule
    \multirow{9}{*}{ProGAN}
     & Wang et al.~\cite{wang2020cnn} & 92.32\,/\,78.23 & 74.29\,/\,57.15 & 61.57\,/\,52.43 & 76.06\,/\,62.60 \\
     & Gragn.\ et al.~\cite{gragnaniello2021gan} & 98.88\,/\,92.83 & 92.86\,/\,64.53 & 72.58\,/\,56.53 & 88.11\,/\,71.30 \\
     & Corvi et al.~\cite{corvi2023detection} & 99.14\,/\,90.67 & 86.06\,/\,57.15 & 66.40\,/\,54.62 & 83.87\,/\,67.48 \\
     & Ojha et al.~\cite{ojha2023universal} & 95.39\,/\,86.13 & 93.66\,/\,82.77 & 75.26\,/\,68.42 & 88.10\,/\,79.11 \\
     & Linear probing~\cite{khan2024clipping} & 98.22\,/\,90.02 & 95.60\,/\,86.01 & 81.95\,/\,72.53 & 91.92\,/\,82.85 \\
     & Fine-tuning~\cite{khan2024clipping} & 99.32\,/\,89.73 & 97.72\,/\,92.59 & 98.52\,/\,94.48 & 98.52\,/\,92.27 \\
     & Adapter~\cite{khan2024clipping} & 99.63\,/\,94.13 & 96.83\,/\,85.70 & 70.29\,/\,55.17 & 88.92\,/\,78.33 \\
     & Prompt tuning~\cite{khan2024clipping} & 99.61\,/\,94.16 & 97.97\,/\,86.86 & 85.71\,/\,59.62 & 94.43\,/\,80.21 \\
    \midrule
    \multirow{5}{*}{\makecell[l]{\dataset{}\\(v0)}}
     & Linear probing & 89.57\,/\,82.97 & 91.68\,/\,82.98 & 88.43\,/\,81.83 & 89.89\,/\,82.59 \\
     & Fine-tuning & 88.99\,/\,84.09 & \textbf{98.35}\,/\,\textbf{94.23} & \textbf{99.90}\,/\,\textbf{98.12} & 95.75\,/\,92.15 \\
     & Adapter & 97.70\,/\,89.76 & 98.02\,/\,92.20 & 99.08\,/\,95.53 & 98.27\,/\,92.50 \\
     & Prompt tuning & 94.74\,/\,87.67 & 97.07\,/\,90.14 & 99.22\,/\,95.62 & 97.01\,/\,91.14 \\
     \rowcolor{ourrow}
     & \method{} & 96.39\,/\,87.49 & 97.04\,/\,91.24 & 98.78\,/\,94.68 & 97.40\,/\,91.14 \\
    \midrule
    \multirow{4}{*}{\makecell[l]{\dataset{}\\(v1)}}
     & Fine-tuning & 87.84\,/\,82.84 & 97.40\,/\,90.93 & 99.88\,/\,94.38 & 95.04\,/\,89.38 \\
     & Adapter & 97.69\,/\,89.75 & 98.02\,/\,92.21 & 99.09\,/\,95.55 & 98.27\,/\,92.50 \\
     & Prompt tuning & 94.74\,/\,87.67 & 97.07\,/\,90.14 & 99.22\,/\,95.62 & 97.01\,/\,91.14 \\
     \rowcolor{ourrow}
     & \method{} & 96.39\,/\,87.49 & 97.04\,/\,91.24 & 98.78\,/\,94.68 & 97.40\,/\,91.14 \\
    \midrule
    \multirow{5}{*}{\makecell[l]{\dataset{}\\(v2)}}
     & Linear probing & 92.17\,/\,84.56 & 92.27\,/\,83.33 & 91.56\,/\,84.38 & 92.00\,/\,84.09 \\
     & Fine-tuning & 88.73\,/\,84.01 & 98.20\,/\,93.94 & 99.82\,/\,97.32 & 95.59\,/\,91.76 \\
     & Adapter & 98.26\,/\,90.06 & 97.86\,/\,87.92 & 99.62\,/\,95.90 & 98.58\,/\,91.29 \\
     & Prompt tuning & 96.41\,/\,87.92 & 96.70\,/\,90.30 & 99.27\,/\,95.25 & 97.46\,/\,91.16 \\
     \rowcolor{ourrow}
     & \method{} & 97.98\,/\,88.33 & 98.33\,/\,93.18 & 99.48\,/\,96.65 & \textbf{98.60}\,/\,\textbf{92.72} \\
    \bottomrule
    \end{tabular}%
    }
\end{table}

\begin{figure}[t]
    \centering
    \includegraphics[width=\linewidth]{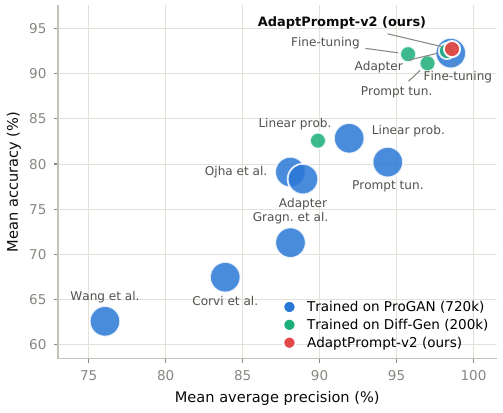}
    \caption{Mean AP versus mean accuracy over the full benchmark (Table~\ref{tab:main}). Bubble area is proportional to training-set size. ProGAN-trained methods (blue) trade off sharply between the two metrics; \dataset{}-trained models (green) cluster in the upper right despite using less than a third of the training data, with \method{}-v2 (red) best on both axes.}
    \label{fig:bubble}
\end{figure}

\subsection{Comparison with Prior Detectors}
Table~\ref{tab:allAP} reports per-dataset AP, Table~\ref{tab:main} aggregates AP and accuracy per generator family, and Fig.~\ref{fig:bubble} plots the overall trade-off.

\paragraph{\textbf{Cross-family generalization:}}
The clearest pattern is the collapse of ProGAN-trained detectors outside the GAN family. \citet{wang2020cnn} falls to 57.15\% accuracy on diffusion test sets and 52.43\% on commercial tools, both close to chance, and even the strong CLIP-based baselines of \citet{khan2024clipping} lose between 13 and 29 AP points when moving from GANs to commercial tools (adapter: 99.63 to 70.29). Training on \dataset{} removes this cliff. \method{}-v2 reaches 98.33\% AP on diffusion models and 99.48\% on commercial tools while still scoring 97.98\% on GANs, and every \dataset{}-trained strategy exceeds its ProGAN-trained counterpart on the non-GAN families. Generalization is asymmetric in a useful direction: models trained on diffusion noise still detect GAN artifacts well, but not vice versa.

\paragraph{\textbf{The aggregate picture:}}
When scores are averaged over the three families, \method{}-v2 attains the best mean AP (98.60\%) and the best mean accuracy (92.72\%) of all 22 configurations, slightly ahead of full CLIP fine-tuning on ProGAN (98.52\% AP, 92.27\% Acc) while training roughly 700$\times$ fewer parameters, and ahead of the strongest single-modality variant on our data (adapter-v2: 98.58\% AP, 91.29\% Acc). The joint visual-textual adaptation is what closes the accuracy gap: adapter-v2 matches \method{}-v2 on AP but trails it by 1.4 accuracy points, indicating better-calibrated decisions when the textual side is adapted as well.

\paragraph{\textbf{Where the weaknesses are:}}
Table~\ref{tab:allAP} is equally clear about what \dataset{} training does not fix. On the face-swap manipulations of FaceForensics++, all \dataset{}-trained models score 55--70 AP, well below the best ProGAN-trained baselines. Face-swap forgeries are partial manipulations of real video frames rather than fully generated images, so their artifacts differ in kind from generative noise; we return to this in the bias analysis below (Section~\ref{sec:bias}). Since the 18-set suite of Table~\ref{tab:allAP} is dominated by GAN test sets and excludes the commercial tools, its mAP column favors ProGAN-trained models; the balanced per-family view of Table~\ref{tab:main} is the fairer summary.

\subsection{Ablation Studies}
\label{sec:ablation}

\begin{figure}[t]
    \centering
    \includegraphics[width=\linewidth]{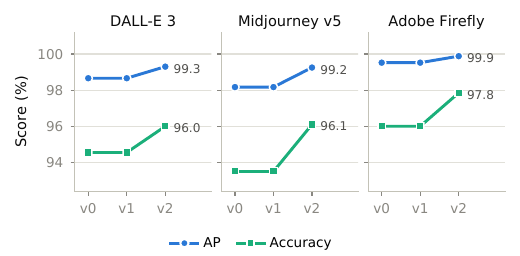}
    \caption{Effect of encoder truncation on the three commercial generators. Removing the last transformer block (v2) improves both AP and accuracy on every tool; removing only the projection layer (v1) changes little.}
    \label{fig:comm-tools}
\end{figure}

\paragraph{\textbf{Truncating the encoder:}}
The v0/v1/v2 blocks of Tables~\ref{tab:allAP} and~\ref{tab:main} isolate the effect of encoder truncation. Removing only the projection layer (v1) leaves adapter- and prompt-based models essentially unchanged, but removing the last transformer block as well (v2) helps consistently: \method{} improves from 97.40\%\,/\,91.14\% (v0) to 98.60\%\,/\,92.72\% (v2) in mean AP\,/\,accuracy, and the adapter baseline shows the same pattern. Fig.~\ref{fig:comm-tools} breaks the comparison down over the three commercial generators, where v2 is uniformly best. These results support the interpretation that CLIP's final block, in service of text-image alignment, smooths away part of the high-frequency structure that separates generated from real images; tapping the penultimate features gives the adapter access to that structure at no extra cost.

\begin{table}[t]
    \caption{Training on ProGAN versus \dataset{} for the four adaptation strategies (v0 encoder; mean AP\,/\,Acc in \%; \dataset{}-trained rows shaded, bold where \dataset{} training wins the pair). \dataset{} training wins everywhere outside the GAN family.}
    \label{tab:diffusionSet}
    \centering
    \resizebox{\linewidth}{!}{%
    \begin{tabular}{llcccc}
    \toprule
    Method & Training & GAN & Diffusion & Comm.\ tools & Average \\
    \midrule
    Linear probing & ProGAN & 98.22\,/\,90.02 & 95.60\,/\,86.01 & 81.95\,/\,72.53 & 91.92\,/\,82.85 \\
     \rowcolor{ourrow}
     & \dataset{} & 89.57\,/\,82.97 & 91.68\,/\,82.98 & \textbf{88.43}\,/\,\textbf{81.83} & 89.89\,/\,82.59 \\
    \midrule
    Fine-tuning & ProGAN & 99.32\,/\,89.73 & 97.72\,/\,92.59 & 98.52\,/\,94.48 & 98.52\,/\,92.27 \\
     \rowcolor{ourrow}
     & \dataset{} & 88.99\,/\,84.09 & \textbf{98.35}\,/\,\textbf{94.23} & \textbf{99.90}\,/\,\textbf{98.12} & 95.75\,/\,92.15 \\
    \midrule
    Adapter & ProGAN & 99.63\,/\,94.13 & 96.83\,/\,85.70 & 70.29\,/\,55.17 & 88.92\,/\,78.33 \\
     \rowcolor{ourrow}
     & \dataset{} & 97.70\,/\,89.76 & \textbf{98.02}\,/\,\textbf{92.20} & \textbf{99.08}\,/\,\textbf{95.53} & \textbf{98.27}\,/\,\textbf{92.50} \\
    \midrule
    Prompt tuning & ProGAN & 99.61\,/\,94.16 & 97.97\,/\,86.86 & 85.71\,/\,59.62 & 94.43\,/\,80.21 \\
     \rowcolor{ourrow}
     & \dataset{} & 94.74\,/\,87.67 & 97.07\,/\,\textbf{90.14} & \textbf{99.22}\,/\,\textbf{95.62} & \textbf{97.01}\,/\,\textbf{91.14} \\
    \bottomrule
    \end{tabular}%
    }
\end{table}

\paragraph{\textbf{Training data:}}
Table~\ref{tab:diffusionSet} pairs each adaptation strategy trained on ProGAN against the same strategy trained on \dataset{}. The pattern is uniform: ProGAN training wins on the GAN family, \dataset{} training wins on diffusion models and, by wide margins, on commercial tools (adapter: 70.29 versus 99.08 AP). For the two parameter-efficient strategies, \dataset{} training also wins on average despite conceding the GAN column, and it does so with 3.6$\times$ less training data. The commercial tools are the deciding column as they are the closest proxy for the generators actually encountered in the wild, and every ProGAN-trained method except full fine-tuning performs poorly on them.

\begin{table}[t]
    \centering
    \caption{Data efficiency: mean AP\,/\,Acc (\%) as the \dataset{} training set shrinks from 80k to 20k images. The ProGAN-trained linear probe of \citet{khan2024clipping} is shown for reference.}
    \label{tab:aptsize}
    \resizebox{\linewidth}{!}{%
    \begin{tabular}{lcccc}
    \toprule
    Method & 20k & 40k & 60k & 80k \\
    \midrule
    Linear probing~\cite{khan2024clipping} & 92.12\,/\,85.49 & 91.26\,/\,85.10 & 91.26\,/\,85.47 & 90.98\,/\,85.06 \\
    \method{}-v0 & 96.61\,/\,90.06 & 97.51\,/\,91.40 & \textbf{97.80\,/\,91.12} & 97.21\,/\,91.37 \\
    \method{}-v2 & \textbf{97.71\,/\,90.38} & \textbf{97.95\,/\,88.87} & 96.40\,/\,85.98 & \textbf{98.04\,/\,91.89} \\
    \bottomrule
    \end{tabular}%
    }
\end{table}

\paragraph{\textbf{Data efficiency and few-shot training:}}
Table~\ref{tab:aptsize} varies the size of the \dataset{} training set. \method{} is remarkably stable: with only 20k images it stays within about one AP point of its 200k-image result, so the benefit of \dataset{} does not depend on scale alone. In a more extreme few-shot setting with only 320 real and 320 fake training images (Table~\ref{tab:fewshot}), the ranking changes: full fine-tuning of the truncated encoder is strongest (97.23\% mean AP), prompt-based variants remain serviceable, and pure adapter variants degrade sharply, suggesting the adapter needs more data than the prompt vectors to find artifact-sensitive directions. \method{}-v2 lands in between at 85.74\%\,/\,77.80\%. We report this negative result deliberately: in data-scarce regimes the best configuration of our framework is not the same as in the full-data regime.

\begin{table}[t]
    \centering
    \caption{Few-shot training with 320 real and 320 fake images (mean AP\,/\,Acc in \%, v0 unless noted).}
    \label{tab:fewshot}
    \resizebox{0.9\linewidth}{!}{%
    \begin{tabular}{llcc}
    \toprule
    Training & Method & Mean AP & Mean Acc \\
    \midrule
    \multirow{4}{*}{ProGAN~\cite{khan2024clipping}}
     & Linear probing & 86.95 & 77.94 \\
     & Fine-tuning & 86.19 & 76.10 \\
     & Adapter & 83.21 & 74.11 \\
     & Prompt tuning & 93.94 & 80.34 \\
    \midrule
    \multirow{6}{*}{\dataset{}}
     & Linear probing & 89.51 & 82.15 \\
     & Fine-tuning & 87.92 & 80.08 \\
     & Fine-tuning (v2) & \textbf{97.23} & \textbf{87.60} \\
     & Adapter & 63.69 & 59.78 \\
     & Prompt tuning & 89.70 & 83.15 \\
     & \method{}-v2 & 85.74 & 77.80 \\
    \bottomrule
    \end{tabular}%
    }
\end{table}

\subsection{Analysis}

\begin{figure}[t]
    \centering
    \includegraphics[width=\linewidth]{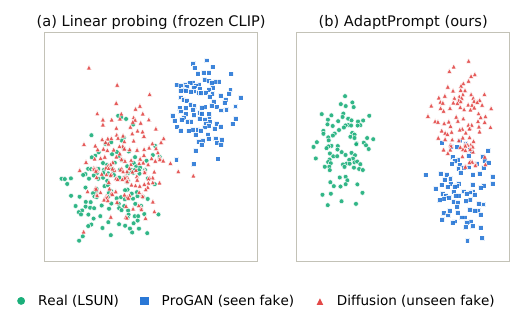}
    \caption{t-SNE of image features. (a)~Frozen CLIP features (linear probing): unseen diffusion fakes overlap the real cluster, the sink-label failure mode. (b)~\method{} features: unseen diffusion fakes join the seen ProGAN fakes in a common synthetic cluster, separated from the real manifold.}
    \label{fig:tsne}
\end{figure}

\paragraph{\textbf{Feature-space view:}}
Fig.~\ref{fig:tsne} visualizes what the adaptation changes. In the frozen CLIP space, unseen diffusion fakes lie inside the real-image cluster providing a geometric picture of the sink-label problem, since no linear boundary can separate them. After adaptation, the same images form a common cluster with the seen ProGAN fakes, clearly separated from the real manifold. The adapter has not memorized a generator; it has moved a family of images it never saw during training onto the synthetic side, which is only possible if the learned directions encode generator-agnostic cues.

\begin{figure}[t]
    \centering
    \includegraphics[width=\linewidth]{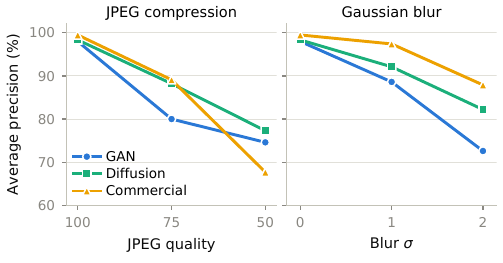}
    \caption{Robustness of \method{}-v2 to post-processing, by generator family. AP at JPEG qualities 100/75/50 (left) and Gaussian blur $\sigma\in\{0,1,2\}$ (right).}
    \label{fig:robustness}
\end{figure}

\paragraph{\textbf{Robustness to post-processing:}}
Images circulating on social platforms are compressed and resampled, which erodes fragile high-frequency evidence. Fig.~\ref{fig:robustness} measures the effect on \method{}-v2. Under JPEG compression at quality 75, AP drops to 80--89\% depending on the family, and to 68--77\% at quality 50; under Gaussian blur, performance degrades more gently (88--97\% at $\sigma{=}1$). Detection of commercial-tool images is the most robust to blur and the most sensitive to heavy compression, consistent with compression destroying precisely the broadband residuals that \dataset{} training teaches the model to use. Robustness under strong compression remains an open problem for this class of detectors.

\paragraph{\textbf{Content bias:}}
\label{sec:bias}
To test whether the detector's accuracy depends on image content rather than provenance, we evaluate on controlled 200+200-image subsets (Table~\ref{tab:bias}). Scene type barely matters: indoor and outdoor images differ by 1.4 AP points. The presence of people matters much more: accuracy drops from 96.5\% to 80.5\% on images containing a person. We see two plausible causes. Human faces dominate the high-quality training data of modern generators, so synthetic people exhibit fewer of the structural defects common in generated objects; and the real class for people spans enormous photometric variation (lighting, occlusion, makeup) that partially mimics the smoothing cues detectors rely on. Together with the weak FaceForensics++ results in Table~\ref{tab:allAP}, this marks person-centric imagery as the clearest limitation of our current model, and face-aware attention is the natural next step.

\begin{table}[t]
    \centering
    \caption{Content-bias probes for \method{}-v2 (\%): 200 real + 200 fake images per condition.}
    \label{tab:bias}
    \begin{tabular}{lcc}
    \toprule
    Subset & AP & Acc \\
    \midrule
    Indoor scenes & 98.47 & 95.00 \\
    Outdoor scenes & 99.91 & 97.00 \\
    Images with people & 97.09 & 80.50 \\
    Images without people & 98.61 & 96.50 \\
    \bottomrule
    \end{tabular}
\end{table}

\begin{figure}[t]
    \centering
    \includegraphics[width=\linewidth]{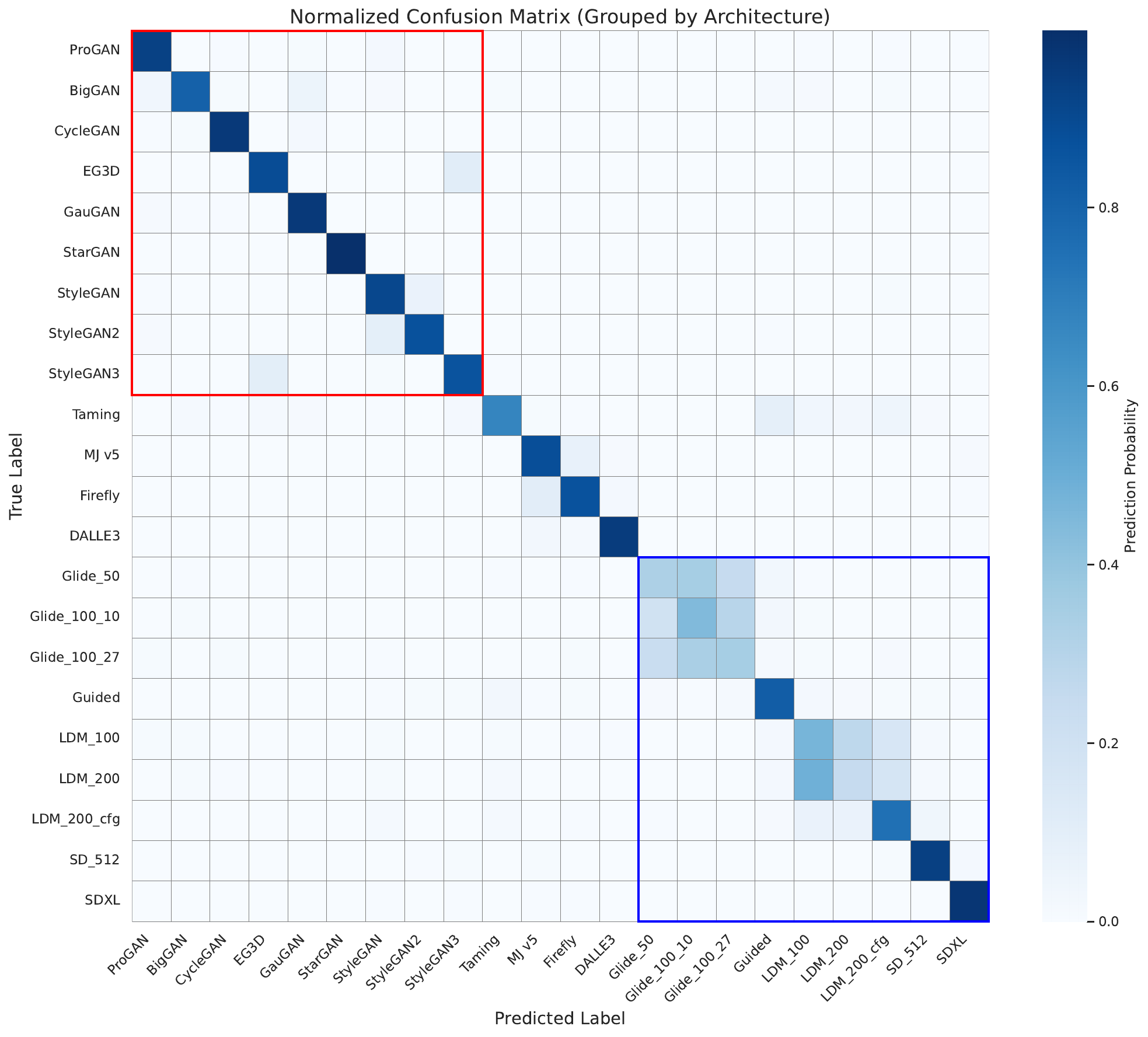}
    \caption{Row-normalized confusion matrix for 22-way source attribution. Red and blue boxes mark the GAN and diffusion families. Errors concentrate inside the boxes---between configurations of the same generator (Glide, LDM)---while the two families remain well separated.}
    \label{fig:confusionMatrix}
\end{figure}

\paragraph{\textbf{Source attribution:}}
Forensic practice often needs to know not just \emph{whether} an image is synthetic but \emph{which} system produced it. We convert \method{} into a 22-way classifier over the generators of our benchmark, training on 640 images per class and testing on the remainder (most test sets contain 360 images per class, so these results are indicative rather than definitive). The model achieves an overall accuracy of 81.4\%, while 96.49\% of its predictions are assigned to the correct generator family. The confusion matrix (Fig.~\ref{fig:confusionMatrix}) shows the expected structure: confusions occur almost exclusively between configurations of the same model, such as the three Glide variants, the three LDM variants, while GANs and diffusion models are almost never confused with one another. Generator fingerprints thus survive the adaptation in enough detail to support attribution, not merely detection.

\section{Conclusion}
We revisited the generalization problem in deepfake detection from the two sides that jointly determine it: what the detector is trained on, and how much of a pre-trained representation the training preserves. On the first, \dataset{} shows that diffusion-generated training data yields detectors that transfer to GANs, diffusion models, and commercial tools alike, whereas the reverse transfer from GAN data fails. On the second, \method{} shows that adapting CLIP's visual and textual pathways jointly, with about 0.1\% of the parameters trainable, matches or exceeds full fine-tuning, and that removing CLIP's final transformer block consistently helps, since the layers that align image features with text also erase forensic detail. The combination attains the best mean AP and accuracy on a 25-set benchmark and extends naturally to source attribution across 22 generators. The limitations are equally concrete: performance drops on person-centric imagery and face-swap manipulations, robustness under strong JPEG compression is limited, and our attribution study is closed-set. Face-aware adaptation, compression-robust training, and open-set attribution are the directions we consider most promising, and we release \dataset{} and our code to support work along these lines.
\begin{acks}
We also acknowledge the use of Generative AI~\cite{openai_chatgpt_2026} for checking and correcting the grammar of this paper. However, it is important to note that we did not use the tool to generate totally new text, rather we use it to check/correct the grammar of our provided text.\\

\noindent \textbf{Author Contributions}
Yichen Jiang conducted the experiments and evaluations, and contributed to the analysis of the results. Mohammed Talha Alam and Sohail Ahmed Khan supervised the experimental work and provided technical guidance throughout the study. Sohail Ahmed Khan led the development and creation of the Diff-Gen dataset. Mohammed Talha Alam contributed to the conceptualization and formulation of the research problem, experimental design, methodological development and supervision, interpretation of the results, and writing and preparation of the manuscript. Duc-Tien Dang-Nguyen and Fakhri Karray served as faculty advisors and provided research guidance and supervision. All authors reviewed and approved the final manuscript.
\end{acks}

%% Uncomment for the camera-ready version.
% \begin{acks}
% Generative AI tools were used to check grammar and phrasing in this
% manuscript; all technical content, experiments, and analyses are the
% authors' own.
% \end{acks}

\bibliographystyle{ACM-Reference-Format}
\bibliography{refs}
% \vspace{-0.38cm}
\newpage
\appendix
\section*{Appendix}
\section{Benchmark Details}
Table~\ref{table:datasets} lists the composition of the 25-set evaluation benchmark: the generator behind each test set, the number of real and fake images, the source of the real images, and the image resolution. Where a test set has no official real counterpart, real images were drawn from LAION to keep the sets balanced.

\begin{table}[H]
    \centering
    \caption{Composition of the evaluation benchmark.}
    \label{table:datasets}
    \resizebox{\linewidth}{!}{%
    \begin{tabular}{lllll}
    \toprule
    Category & Generator & Real\,/\,fake & Real source & Resolution \\
    \midrule
    \multirow{10}{*}{GAN}
      & ProGAN~\cite{karras2018progressive} & 4k\,/\,4k & LSUN & $256^2$ \\
      & BigGAN~\cite{Brock2018LargeSG} & 2k\,/\,2k & ImageNet & $256^2$ \\
      & CycleGAN~\cite{zhu2017cyclegan} & 1k\,/\,1k & Various & $256^2$ \\
      & EG3D~\cite{chan2022efficient} & 1k\,/\,1k & LAION & $512^2$ \\
      & GauGAN~\cite{park2019semantic} & 5k\,/\,5k & COCO & $256^2$ \\
      & StarGAN~\cite{choi2018stargan} & 2k\,/\,2k & CelebA & $256^2$ \\
      & StyleGAN~\cite{karras2019stylegan} & 1k\,/\,1k & LSUN & $256^2$ \\
      & StyleGAN2~\cite{karras2020stylegan2} & 1k\,/\,1k & Various & ${\approx}256^2$ \\
      & StyleGAN3~\cite{karras2021stylegan3} & ${\approx}$1k\,/\,1k & Various & $512^2$ \\
      & Taming Transf.~\cite{esser2021taming} & 1k\,/\,1k & LAION & $256^2$ \\
    \midrule
    \multirow{9}{*}{Diffusion}
      & Glide (50/27)~\cite{nichol2021glide} & 1k\,/\,1k & LAION & $256^2$ \\
      & Glide (100/10)~\cite{nichol2021glide} & 1k\,/\,1k & LAION & $256^2$ \\
      & Glide (100/27)~\cite{nichol2021glide} & 1k\,/\,1k & LAION & $256^2$ \\
      & Guided Diffusion~\cite{openai2021guideddiffusion} & 1k\,/\,1k & LAION & $256^2$ \\
      & LDM (100)~\cite{rombach2022latentdiffusion} & 1k\,/\,1k & LAION & $256^2$ \\
      & LDM (200)~\cite{rombach2022latentdiffusion} & 1k\,/\,1k & LAION & $256^2$ \\
      & LDM (200, CFG)~\cite{rombach2022latentdiffusion} & 1k\,/\,1k & LAION & $256^2$ \\
      & Stable Diffusion~\cite{rombach2022latentdiffusion} & 1k\,/\,1k & LAION & $512^2$ \\
      & SDXL~\cite{podell2023sdxl} & 1k\,/\,1k & LAION & $1024^2$ \\
    \midrule
    \multirow{3}{*}{Comm.\ tools}
      & Midjourney v5~\cite{midjourney2023} & 1k\,/\,1k & LAION & Various \\
      & Adobe Firefly~\cite{adobe2023firefly} & 1k\,/\,1k & LAION & Various \\
      & DALL-E 3~\cite{openai2023dalle3} & 1k\,/\,1k & LAION & Various \\
    \midrule
    \multirow{3}{*}{Other}
      & DALL-E mini~\cite{dayma2021dallemini} & 1k\,/\,1k & LAION & $256^2$ \\
      & DeepFakes~\cite{rossler2019faceforensics} & ${\approx}$2.7k\,/\,2.7k & YouTube & ${\approx}256^2$ \\
      & FaceSwap~\cite{rossler2019faceforensics} & 2.8k\,/\,2.8k & YouTube & ${\approx}256^2$ \\
    \bottomrule
    \end{tabular}%
    }
\end{table}

\begin{figure}[htp]
    \centering
    \includegraphics[width=\linewidth]{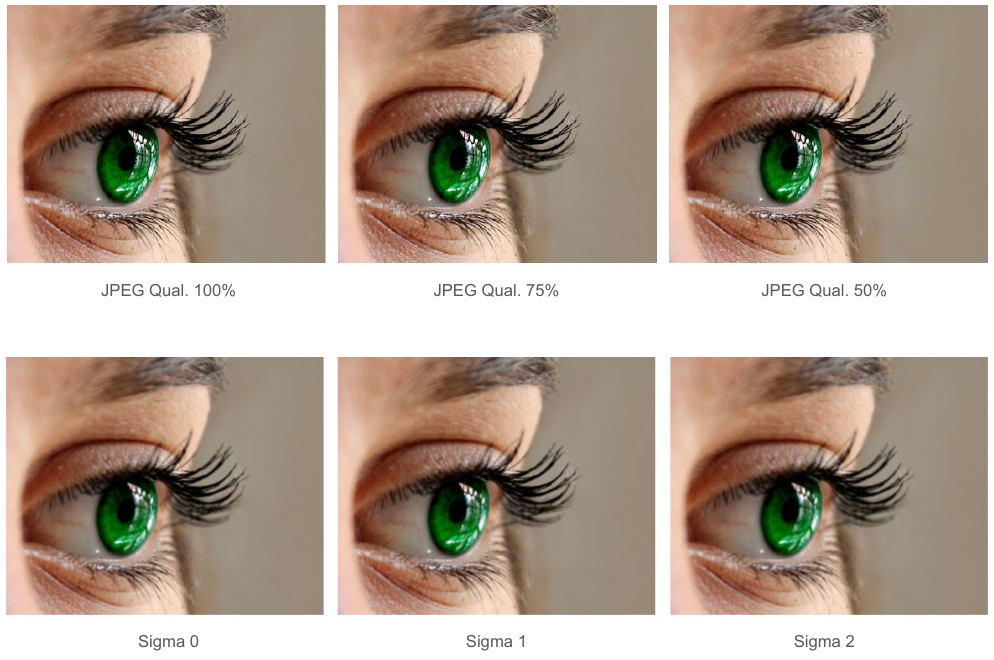}
    \caption{Qualitative examples of the post-processing perturbations used in the robustness evaluation. The top row shows JPEG compression at qualities 100, 75, and 50, while the bottom row shows Gaussian blur with $\sigma \in \{0,1,2\}$. Quantitative robustness results are reported in Fig.~\ref{fig:robustness}.}
    \label{fig:robustness2}
\end{figure}

\section{Implementation Details}
\label{app:implementation}

Table~\ref{tab:implementation} summarizes the main configuration of
AdaptPrompt. All CLIP backbone parameters remain frozen during training,
and only the visual adapter and learnable textual context vectors are
optimized. The same configuration is used across experiments unless
stated otherwise.

\begin{table}[H]
\centering
\caption{Implementation details of AdaptPrompt.}
\label{tab:implementation}
\small
\begin{tabular}{ll}
\toprule
\textbf{Component} & \textbf{Configuration} \\
\midrule
Backbone & CLIP ViT-L/14 \\
Backbone parameters & 427M \\
Joint embedding dimension & 768 \\
Visual adapter & $768 \rightarrow 384 \rightarrow 768$ \\
Adapter activation & ReLU \\
Visual adapter parameters & $\sim$590k \\
Prompt context length & 16 vectors \\
Prompt parameters & $\sim$12k \\
Total trainable parameters & $\sim$602k \\
Trainable fraction & $\sim$0.1\% \\
Backbone weights & Frozen \\
Training objective & Cross-entropy \\
Prediction & Cosine similarity \\
\bottomrule
\end{tabular}
\end{table}

\paragraph{\textbf{Encoder variants:}}
We evaluate three configurations of the CLIP vision encoder. The
\textbf{v0} configuration uses the complete pre-trained backbone;
\textbf{v1} removes the final visual projection layer; and \textbf{v2}
additionally removes the last transformer block, allowing the adapter
to operate on the penultimate-block representation. All remaining CLIP
parameters stay frozen. Among these configurations, v2 consistently
provides the strongest cross-generator performance.

\paragraph{\textbf{Inference protocol:}}
At inference time, AdaptPrompt requires only the input image and the
learned real/fake textual prompts. No generator-specific calibration,
post-processing, or access to the source generative model is required.
Predictions are obtained directly from the cosine similarities between
the adapted visual representation and the two class embeddings. This
keeps the deployment procedure identical across all unseen generators
and avoids introducing test-set-specific adjustments.

\paragraph{\textbf{Evaluation protocol:}}
Unless explicitly stated otherwise, all evaluations are cross-generator,
so no test-set generator is observed during training. We report average
precision (AP) together with classification accuracy at a fixed threshold
of 0.5. For generators represented by multiple configurations, such as
Glide and LDM, scores are averaged across their respective variants.

\begin{table*}[t]
  \centering
  \caption{Classification accuracy on unseen test sets (all values \%; fixed 0.5 threshold). Layout follows Table~\ref{tab:allAP}: bold marks columns where a \dataset{}-trained model is best overall; shaded rows are \method{}. Glide and LDM scores are averaged over their three configurations each.}
  \label{tab:allAcc}
  \resizebox{\textwidth}{!}{%
    \begin{tabular}{llccccccccccccccccccc}
    \toprule
    \multirow{2}{*}{Training} & \multirow{2}{*}{Method}
      & \multicolumn{10}{c}{Generative adversarial networks}
      & \multirow{2}{*}{\makecell{DALL-E\\(mini)}}
      & \multicolumn{5}{c}{Diffusion models}
      & \multicolumn{2}{c}{FaceForensics++}
      & \multirow{2}{*}{\makecell{Avg.\\Acc}} \\
    \cmidrule(lr){3-12} \cmidrule(lr){14-18} \cmidrule(lr){19-20}
      & & \makecell{Pro\\GAN} & \makecell{Big\\GAN} & \makecell{Cycle\\GAN} & EG3D & \makecell{Gau\\GAN} & \makecell{Star\\GAN}
      & \makecell{Style\\GAN} & \makecell{Style\\GAN2} & \makecell{Style\\GAN3} & \makecell{Taming\\Transf.}
      & & Glide & Guided & LDM & SD & SDXL
      & \makecell{Deep\\Fakes} & \makecell{Face\\Swap} & \\
    \midrule
    \multirow{10}{*}{ProGAN}
      & Wang et al.~\cite{wang2020cnn} (B+J 0.1)
      & 99.90 & 67.65 & 79.50 & 72.65 & 76.63 & 89.72 & 82.10 & 77.05 & 80.68 & 56.45 & 55.05 & 61.15 & 62.90 & 54.03 & 52.50 & 53.40 & 52.67 & 49.68 & 66.09 \\
      & Wang et al.~\cite{wang2020cnn} (B+J 0.5)
      & 99.65 & 58.13 & 77.80 & 50.30 & 75.56 & 79.99 & 69.80 & 62.30 & 53.42 & 51.05 & 51.90 & 54.33 & 52.35 & 51.35 & 50.15 & 51.00 & 51.46 & 50.02 & 59.18 \\
      & Gragnaniello et al.~\cite{gragnaniello2021gan}
      & 100.00 & 93.27 & 91.75 & 97.55 & 94.13 & 99.65 & 97.25 & 89.75 & 97.47 & 67.45 & 60.65 & 69.38 & 67.30 & 62.33 & 59.70 & 57.75 & 65.31 & 50.02 & 76.59 \\
      & Corvi et al.~\cite{corvi2023detection} (ProGAN)
      & 100.00 & 95.85 & 90.35 & 98.40 & 92.46 & 99.00 & 97.65 & 84.90 & 82.79 & 65.30 & 69.30 & 58.98 & 53.10 & 58.83 & 55.70 & 52.10 & 59.38 & 50.11 & 72.72 \\
      & Corvi et al.~\cite{corvi2023detection} (Latent)
      & 50.94 & 51.82 & 46.20 & 49.25 & 50.86 & 48.02 & 59.40 & 50.95 & 50.05 & 77.65 & 87.00 & 59.83 & 50.95 & 99.25 & 99.25 & 93.10 & 69.87 & 48.14 & 66.40 \\
      & Ojha et al.~\cite{ojha2023universal}
      & 98.94 & 94.48 & 94.20 & 57.75 & 94.65 & 87.49 & 85.55 & 83.40 & 75.42 & 89.45 & 89.20 & 82.15 & 79.00 & 87.80 & 81.90 & 74.15 & 62.71 & 64.30 & 82.84 \\
      & Linear probing~\cite{khan2024clipping}
      & 98.50 & 91.75 & 91.00 & 98.20 & 88.08 & 94.42 & 81.40 & 71.70 & 94.11 & 91.05 & 85.80 & 90.55 & 79.05 & 87.42 & 77.30 & 83.85 & 69.37 & 68.30 & 86.26 \\
      & Fine-tuning~\cite{khan2024clipping}
      & 99.60 & 77.38 & 71.55 & 98.40 & 65.70 & 100.00 & 94.85 & 95.30 & 99.89 & 94.40 & 93.20 & 88.78 & 92.35 & 95.17 & 91.75 & 97.35 & 76.46 & 52.11 & 88.74 \\
      & Adapter~\cite{khan2024clipping}
      & 99.88 & 94.75 & 97.45 & 95.30 & 95.47 & 99.12 & 93.35 & 78.35 & 93.11 & 94.55 & 92.00 & 94.27 & 81.65 & 89.18 & 67.70 & 71.60 & 77.11 & 70.16 & 88.72 \\
      & Prompt tuning~\cite{khan2024clipping}
      & 99.83 & 93.80 & 95.60 & 93.50 & 93.43 & 99.15 & 95.25 & 82.95 & 93.11 & 94.95 & 91.50 & 92.88 & 84.30 & 88.16 & 76.45 & 77.80 & 78.45 & 74.66 & 89.45 \\
    \midrule
    \multirow{5}{*}{\makecell[l]{\dataset{}\\(v0)}}
      & Linear probing
      & 89.14 & 64.95 & 66.85 & 95.10 & 60.13 & 79.83 & 95.15 & 86.70 & 97.58 & 94.15 & 92.65 & 71.57 & 89.45 & 84.07 & 95.25 & 95.25 & 49.02 & 51.04 & 80.59 \\
      & Fine-tuning
      & 88.79 & 65.12 & 63.50 & 98.50 & 53.78 & 96.82 & 87.25 & 89.80 & 99.68 & \textbf{97.65} & \textbf{97.70} & 87.55 & \textbf{94.50} & 98.05 & 98.40 & \textbf{98.35} & 52.43 & 51.21 & 87.39 \\
      & Adapter
      & 98.24 & 73.95 & 79.40 & 98.35 & 74.52 & 95.42 & 95.20 & 86.75 & 98.89 & 96.85 & 96.45 & 88.68 & 91.00 & 92.18 & 98.35 & 97.85 & 54.45 & 57.11 & 88.88 \\
      & Prompt tuning
      & 93.76 & 73.65 & 73.55 & 97.60 & 72.09 & 90.40 & 95.05 & 86.30 & 98.84 & 95.50 & 95.65 & 84.98 & 89.85 & 90.62 & 97.50 & 97.10 & 51.58 & 52.00 & 86.96 \\
      \rowcolor{ourrow}
      & \method{}
      & 98.40 & 69.72 & 71.35 & 96.20 & 66.65 & 93.87 & 95.35 & 91.00 & 97.84 & 94.55 & 94.65 & 88.18 & 89.20 & 91.75 & 96.20 & 96.00 & 53.63 & 54.95 & 87.34 \\
    \midrule
    \multirow{4}{*}{\makecell[l]{\dataset{}\\(v1)}}
      & Fine-tuning
      & 92.51 & 61.23 & 61.10 & 94.55 & 51.83 & 96.17 & 88.75 & 90.40 & 98.05 & 93.85 & 93.05 & 85.23 & 90.60 & 94.28 & 94.55 & 94.55 & 50.19 & 51.32 & 84.98 \\
      & Adapter
      & 98.24 & 73.95 & 79.35 & 98.35 & 74.50 & 95.42 & 95.20 & 86.75 & 98.89 & 96.85 & 96.45 & 88.71 & 91.00 & 92.20 & 98.35 & 97.85 & 54.45 & 57.11 & 88.88 \\
      & Prompt tuning
      & 93.76 & 73.65 & 73.55 & 97.60 & 72.09 & 90.40 & 95.05 & 86.30 & 98.84 & 95.50 & 95.65 & 84.98 & 89.85 & 90.62 & 97.50 & 97.10 & 51.58 & 52.00 & 86.96 \\
      \rowcolor{ourrow}
      & \method{}
      & 98.40 & 69.72 & 71.35 & 96.20 & 66.65 & 93.87 & 95.35 & 91.00 & 97.84 & 94.55 & 94.65 & 88.18 & 89.20 & 91.75 & 96.20 & 96.00 & 53.63 & 54.95 & 87.34 \\
    \midrule
    \multirow{5}{*}{\makecell[l]{\dataset{}\\(v2)}}
      & Linear probing
      & 93.24 & 68.58 & 65.50 & 93.95 & 64.83 & 87.97 & 93.70 & 86.70 & 96.37 & 94.80 & \textbf{97.70} & 72.88 & 89.05 & 83.52 & 95.90 & 95.80 & 52.43 & 51.21 & 82.00 \\
      & Fine-tuning
      & 92.11 & 64.78 & 65.55 & 98.10 & 53.57 & 94.20 & 87.70 & 87.75 & 99.26 & 97.10 & 96.75 & 88.23 & 92.95 & 97.32 & 98.10 & 97.80 & 50.73 & 52.00 & 87.08 \\
      & Adapter
      & 97.45 & 80.67 & 80.90 & \textbf{99.10} & 79.98 & 97.72 & 90.40 & 79.45 & 99.05 & 95.85 & 93.50 & 81.63 & 85.75 & 87.73 & 99.15 & 98.25 & 55.00 & 54.82 & 87.31 \\
      & Prompt tuning
      & 98.20 & 70.15 & 72.90 & 96.15 & 64.22 & 97.85 & 95.20 & 90.40 & 99.85 & 94.25 & 93.65 & 87.05 & 88.90 & 90.27 & 96.15 & 95.70 & 52.30 & 54.12 & 87.11 \\
      \rowcolor{ourrow}
      & \method{}
      & 98.46 & 70.47 & 70.10 & 98.10 & 69.99 & 97.72 & 94.35 & 89.80 & 97.26 & 97.00 & 95.75 & 90.70 & 93.50 & 92.33 & 98.10 & 97.90 & 57.51 & 58.18 & 88.93 \\
    \bottomrule
    \end{tabular}%
  }
\end{table*}

\section{Per-Dataset Accuracy}
\label{sec:appacc}
Table~\ref{tab:allAcc} complements Table~\ref{tab:allAP} with per-dataset classification accuracy at a fixed 0.5 threshold. The trends mirror the AP results: ProGAN-trained models dominate the GAN columns, while \dataset{}-trained models are markedly stronger on the diffusion test sets, DALL-E~mini, and (per Table~\ref{tab:main}) the commercial tools, which this suite does not include. Accuracy is more sensitive than AP to threshold calibration, which explains the larger gaps between the two metrics for the baselines.
% \newpage
% \section{Ethical Considerations}
% Detection research in this space is inherently dual-use: publishing which cues a detector relies on can, in principle, help an adversary suppress them. We consider the balance clearly in favor of disclosure --defenders, platforms, and forensic practitioners benefit from knowing both the strengths and the failure modes documented here (person-centric imagery, heavy recompression), and counter-forensic adaptation is possible with or without publication. \dataset{} contains no personal data beyond what is already present in the public LSUN corpus; its synthetic half is generated from generic object prompts and depicts no identifiable individuals. The evaluation reuses established public benchmarks under their original licenses. We release \dataset{} and our code for research use, and we caution that scores reported under laboratory conditions, including ours, should not be treated as reliable evidence thresholds in legal or journalistic settings without case-specific validation.

\end{document}